\documentclass{llncs}

\usepackage[british]{babel}
\usepackage{graphicx}

\usepackage{paralist}

\title{Dominion}\subtitle{A constraint solver generator\vspace*{-.5em}}
\author{Lars Kotthof\/f\\\email{larsko@cs.st-andrews.ac.uk}\vspace*{-.5em}}
\institute{University of St Andrews}

\begin{document}
\maketitle
\vspace*{-1.5em}
\begin{abstract}
This paper proposes a design for a system to generate constraint solvers that
are specialised for specific problem models. It describes the design in detail
and gives preliminary experimental results showing the feasibility and
effectiveness of the approach.
\end{abstract}

\vspace*{-1.7em}
\section{Introduction}

\noindent Currently, applying constraint technology to a large, complex problem
requires significant manual tuning by an expert. Such experts are rare.  The
central aim of this project is to improve the scalability of constraint
technology, while simultaneously removing its reliance on manual tuning by an
expert.  We propose a novel, elegant means to achieve this -- a \emph{constraint
solver synthesiser}, which generates a constraint solver specialised to a given
problem. Constraints research has mostly focused on the incremental improvement
of general-purpose solvers so far. The closest point of comparison is currently
the G12 project~\cite{g12}, which aims to combine existing general constraint
solvers and solvers from related fields into a hybrid. There are previous
efforts at generating specialised constraint solvers in the literature,
e.g.~\cite{minton}; we aim to use state-of-the-art constraint solver technology
employing a broad range of different techniques. Synthesising a constraint
solver has two key benefits. First, it will enable a fine-grained optimisation
not possible for a general solver, allowing the solving of much larger, more
difficult problems.  Second, it will open up many new research possibilities.
There are many techniques in the literature that, although effective in a
limited number of cases, are not suitable for general use. Hence, they are
omitted from current general solvers and remain relatively undeveloped.  Among
these are for example conflict recording~\cite{nogoods}, backjumping~\cite{cbj},
singleton arc consistency~\cite{sac}, and neighbourhood inverse
consistency~\cite{inversecons}. The synthesiser will select such techniques as
they are appropriate for an input problem. Additionally, it can also vary basic
design decisions, which can have a significant impact on
performance~\cite{survey}.

\smallskip

The system we are proposing in this paper, Dominion, implements a design that is
capable of achieving said goals effectively and efficiently. The design
decisions we have made are based on our experience with Minion~\cite{minion} and
other constraint programming systems.

\smallskip

The remainder of this paper is structured as follows. In the next section, we
describe the design of Dominion and which challenges it addresses in particular.
We then present the current partial implementation of the proposed system and
give experimental results obtained with it. We conclude by proposing directions
for future work.

\section{Design of a synthesiser for specialised constraint solvers}

\begin{figure}[bt]
\begin{center}
\includegraphics{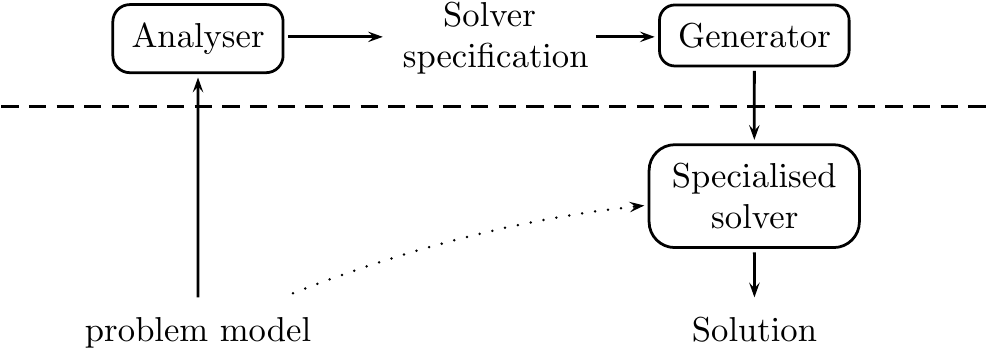}
\end{center}
\caption{Components and flow of information in Dominion. The part above the
dashed line is the actual Dominion system. The dotted arrow from the problem
model to the specialised solver designates that either the model is encoded
entirely in the solver such that no further information is required to solve the
problem, or the solver requires further input such as problem
parameters.\label{design}}
\end{figure}

\noindent The design of Dominion distinguishes two main parts. The
\emph{analyser} analyses the problem model and produces a solver specification
that describes what components the specialised solver needs to have and which
algorithms and data structures to use. The \emph{generator} takes the solver
specification and generates a solver that conforms to it. The flow of
information is illustrated in Figure~\ref{design}.

Both the analyser and the generator optimise the solver. While the analyser
performs the high-level optimisations that depend on the structure of the
problem model, the generator performs low-level optimisations which depend on
the implementation of the solver. Those two parts are independent and linked by
the solver specification, which is completely agnostic of the format of the
problem model and the implementation of the specialised solver. There can be
different front ends for both the analyser and the generator to handle problems
specified in a variety of formats and specialise solvers in a number of
different ways, e.g.\ based on existing building blocks or synthesised
from scratch.

\subsection{The analyser}

\noindent The analyser operates on the model of a constraint problem class or
instance. It determines the constraints, variables, and associated domains
required to solve the problem and reasons about the algorithms and data
structures the specialised solver should use. It makes high-level design
decisions, such as whether to use trailing or copying for backtracking memory.
It also decides what propagation algorithms to use for specific constraints and
what level of consistency to enforce.

The output of the analyser is a solver specification that describes all the
design decisions made. It does not necessarily fix all design decisions -- it
may use default values -- if the analyser is unable to specialise a particular
part of the solver for a particular problem model.

In general terms, the requirements for the solver specification are that it
\begin{inparaenum}[(a)]
\item describes a solver which is able to find solutions to the analysed problem
    model and
\item describes optimisations which will make this solver perform better than a
    general solver.
\end{inparaenum}

The notion of better performance includes run time as well as other resources
such as memory. It is furthermore possible to optimise with respect to a
particular resource; for example a solver which uses less memory at the expense
of run time for embedded systems with little memory can be specified.

The solver specification may include a representation of the original problem
model such that a specialised solver which encodes the problem can be produced
-- the generated solver does not require any input when run or only values
for the parameters of a problem class. It may furthermore modify the original
model in a limited way; for example split variables which were defined as one
type into several new types. It does not, however, optimise it like for example
Tailor~\cite{tailor}.

The analyser may read a partial solver specification along with the model of the
problem to be analysed to still allow fine-tuning by human experts while not
requiring it. This also allows for running the analyser incrementally, refining
the solver specification based on analysis and decisions made in earlier steps.

\smallskip

The analyser creates a constraint optimisation model of the problem of
specialising a constraint solver. The decision variables are the design
decisions to be made and the values in their domains are the options which are
available for their implementation. The constraints encode which parts are
required to solve the problem and how they interact. For example, the
constraints could require the presence of an integer variable type and an equals
constraint which is able to handle integer variables. A solution to this
constraint problem is a solver specification that describes a solver which is
able to solve the problem described in the original model. The weight attached
to each solution describes the performance of the specialised solver and could
be based on static measures of performance as well as dynamic ones; e.g.\
predefined numbers describing the performance of a specific algorithm and
experimental results from probing a specific implementation.

This metamodel enables the use of constraint programming techniques for
generating the specialised solver and ensures that a solver specification can be
created efficiently even for large metamodels.

\smallskip

The result of running the analyser phase of the system is a solver specification
which specifies a solver tailored to the analysed problem model.

\subsection{The generator}

\noindent The generator reads the solver specification produced by the analyser
and constructs a specialised constraint solver accordingly. It may modify an
existing solver, or synthesise one from scratch. The generated solver has to
conform to the solver specification, but beyond that, no restrictions are
imposed. In particular, the generator does not guarantee that the generated
specialised solver will have better performance than a general solver, or indeed
be able to solve constraint problems at all -- this is encoded in the solver
specification.

In addition to the high-level design decisions fixed in the solver
specification, the generator can perform low-level optimisations which are
specific to the implementation of the specialised solver. It could for example
decide to represent domains with a data type of smaller range than the default
one to save space.

The scope of the generator is not limited to generating the source code
which implements the specialised solver, but also includes the system to build
it.

The result of running the generator phase of the system is a specialised solver
which conforms to the solver specification.

\section{Preliminary implementation and experimental results}

\noindent We have started implementing the design proposed above in a system
which operates on top of Minion~\cite{minion}. The analyser reads Minion input
files and writes a solver specification which describes the constraints and the
variable types which are required to solve the problem. It does not currently
create a metamodel of the problem. The generator modifies Minion to support only
those constraints and variable types. It furthermore does some additional
low-level optimisations by removing infrastructure code which is not required
for the specialised solver. The current implementation of Dominion sits between
the existing Tailor and Minion projects -- it takes Minion problem files, which
may have been generated by Tailor, as input, and generates a specialised Minion
solver.

The generated solver is specialised for models of problem instances from the
problem class the analysed instance belongs to. The models have to be the same
with respect to the constraints and variable types used.

\begin{figure}[!tb]
\includegraphics{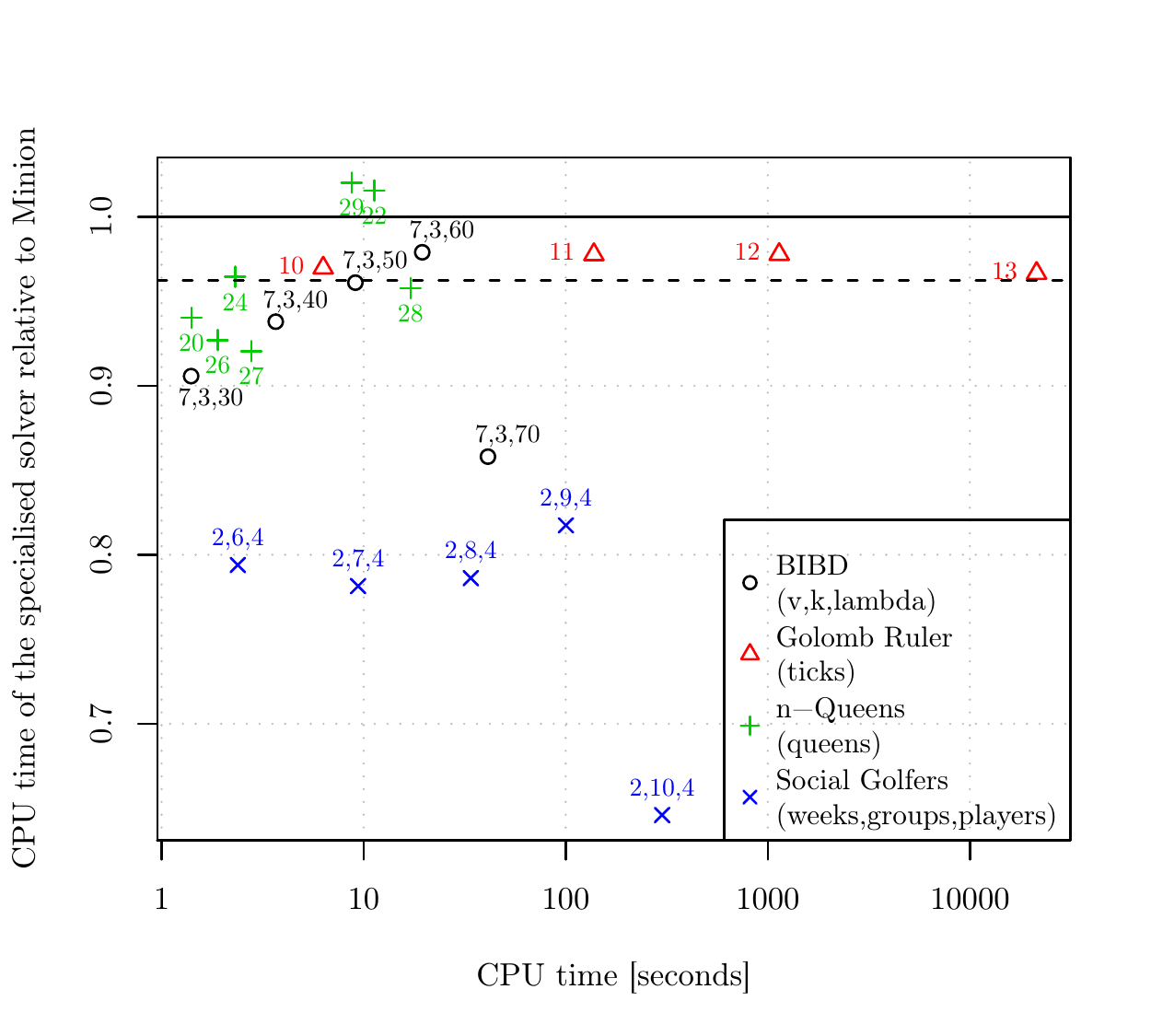}
\vspace*{-2em}
\caption{Preliminary experimental results for models of instances of four
problem classes. The $x$ axis shows the time standard Minion took to solve the
respective instance. The labels of the data points show the parameters of the
problem instance, which are given in parentheses in the legend. The times
were obtained using a development version of Minion which corresponds to release
0.8.1 and Dominion-generated specialised solvers based on the same version of
Minion. Symbols below the solid line designate problem instances where the
Dominion-generated solver was faster than Minion. The points above the line are
not statistically significant; they are random noise. The dashed line designates
the median for all problem instances.\label{results}}
\end{figure}

Experimental results for models from four different problem classes are shown
in Figure~\ref{results}. The graph only compares the CPU time Minion and the
specialised solver took to solve the problem; it does not take into account the
overhead of running Dominion -- analysing the problem model, generating the
solver, and compiling it, which was in the order of a few minutes for all of the
benchmarks.

The problem classes Balanced Incomplete Block Design, Golomb Ruler, $n$-Queens,
and Social Golfers were chosen because they use a range of different
constraints and variable types. Hence the optimisations Dominion can perform are
different for each of these problem classes. This is reflected in the
experimental results by different performance improvements for different
classes.

Figure~\ref{results} illustrates two key points. The first point is that even a
quite basic implementation of Dominion which does only a few optimisations can
yield significant performance improvements over standard Minion. The second
point is that the performance improvement does not only depend on the problem
class, but also on the instance, even if no additional optimisations beyond the
class level were performed. For both the Balanced Incomplete Block Design and
the Social Golfers problem classes the largest instances yield significantly
higher improvements than smaller ones.

At this stage of the implementation, our aim is to show that a specialised
solver can perform better than a general one. We believe that
Figure~\ref{results} conclusively shows that. As the problem models
become larger and take longer to solve, the improvement in terms of absolute run
time difference becomes larger as well. Hence the more or less constant overhead
of running Dominion is amortised for larger and more difficult problem models,
which are our main focus. Generating a specialised solver for problem classes
and instances is always going to entail a certain overhead, making the approach
infeasible for small and quick-to-solve problems.

\section{Conclusion and future work}

\noindent We have described the design of Dominion, a solver generator, and
demonstrated its feasibility by providing a preliminary implementation. We have
furthermore demonstrated the feasibility and effectiveness of the general
approach of generating specialised constraint solvers for problem models by
running experiments with Minion and Dominion-generated solvers and obtaining
results which show significant performance improvements. These results do not
take the overhead of running Dominion into account, but we are confident that
for large problem models there will be an overall performance improvement
despite the overhead.

Based on our experiences with Dominion, we propose that the next step should be
the generation of specialised variable types for the model of a problem
instance. Dominion will extend Minion and create variable types of the sort
``Integer domain ranging from 10 to 22''. This not only allows us to choose
different representations for variables based on the domain, but also to
simplify and speed up services provided by the variable, such as checking the
bounds of the domain or checking whether a particular value is in the domain.

The implementation of specialised variable types requires generating solvers for
models of problem instances because the analysed problem model is essentially
rewritten. The instance the solver was specialised for will be encoded in it and
no further input will be required to solve the problem.  We expect this
optimisation to provide an additional improvement in performance which is more
consistent across different problem classes, i.e.\ we expect significant
improvements for all problem models and not just some.

We are also planning on continuing to specify the details of Dominion and
implementing it.

\section{Acknowledgements}

\noindent The authors thank Chris Jefferson for extensive help with the
internals of Minion and the anonymous reviewers for their feedback. Lars
Kotthof\/f is supported by a SICSA studentship.

\end{document}